%% file: main.tex
\documentclass[conference,10pt]{IEEEtran}
\usepackage{geometry}
\usepackage{algpseudocode}
\algtext*{EndWhile} 
\algtext*{EndIf}
\algtext*{EndFor}
\algtext*{EndProcedure}
\usepackage{amsmath}
\usepackage{amssymb}
\usepackage{array}
\usepackage{algorithm}
\usepackage{bm}
\usepackage[bookmarks=false]{hyperref}
\usepackage[noadjust]{cite}
\usepackage[font=small]{caption}
\usepackage{color}
\usepackage{comment}
\usepackage{dblfloatfix}
\usepackage[inline]{enumitem}
\usepackage{epsfig}
\usepackage{etoolbox}
\usepackage{fancybox}
\usepackage{float}
\usepackage{fixltx2e}
\usepackage{graphicx}
\usepackage{hyperref}
\usepackage{listings}
\usepackage{mathrsfs}
\usepackage{multirow}
\usepackage{multicol}
\usepackage{pifont}
\usepackage{placeins}
\usepackage{rotating}
\usepackage{setspace}
\usepackage{soul}
\usepackage[hang, tight]{subfigure}
\usepackage{threeparttable}
\usepackage{times}
\usepackage{url}
\usepackage{upgreek}
\usepackage{verbatim}
\usepackage{wrapfig}
\usepackage[usenames,dvipsnames]{xcolor}
\usepackage[]{siunitx}
\usepackage[super]{nth}
\usepackage{tabularx}
\usepackage{supertabular}
\usepackage{tikz}
\usepackage{bbm}
\usepackage{blindtext}

\newtheorem{lemma}{\bf Lemma}

\graphicspath{{./_fig/}}

\begin{document}
\title{\huge{JMSNAS: Joint Model Split and Neural Architecture Search for Learning over Mobile Edge Networks}
\vspace{-2mm}}
\author{\IEEEauthorblockN{Yuqing Tian,
Zhaoyang Zhang$^\dag$, Zhaohui Yang, and Qianqian Yang}
\IEEEauthorblockA{College of Information Science and Electronic Engineering, Zhejiang University, Hangzhou, China\\
Zhejiang Provincial Key Laboratory of Info. Proc., Commun. \& Netw. (IPCAN), Hangzhou, China\\
E-mail: \{tianyq, ning\_ming$^\dag$\}@zju.edu.cn, zhaohui.yang@ucl.ac.uk, qianqianyang20@zju.edu.cn
}
}

\maketitle
\begingroup\renewcommand\thefootnote{*}
\footnotetext{This work was supported in part by National Key R\&D Program of China under Grant 2020YFB1807101, and National Natural Science Foundation of China under Grant U20A20158 and 61725104.}
\endgroup

\input{_txt/abstract}

\input{_txt/1_introduction}

\input{_txt/2_preliminaries}
\input{_txt/3_Problem_Formulation}

\input{_txt/4_Methodology}

\input{_txt/5_experiment}

\input{_txt/6_conclusion}

\bibliographystyle{IEEEtran}
\bibliography{references} 
\end{document}

%% file: _txt/abstract.tex
\begin{abstract}

    The main challenge to deploy deep neural network (DNN) over a mobile edge network is how to split the DNN model so as to match the network architecture as well as all the nodes' computation and communication capacity. This essentially involves two highly coupled procedures: model generating and model splitting.  In this paper, a joint model split and neural architecture search (JMSNAS) framework is proposed to automatically generate and deploy a DNN model over a mobile edge network. Considering both the computing and communication resource constraints, a computational graph search problem is formulated to find the multi-split points of the DNN model, and then the model is trained to meet some accuracy requirements. Moreover, the trade-off between model accuracy and completion latency is achieved through the proper design of the objective function. The experiment results confirm the superiority of the proposed framework over the state-of-the-art split machine learning design methods. 
\end{abstract}

%% file: _txt/1_introduction.tex
\section{Introduction}
The fifth-generation (5G) mobile networks are envisioned to support the booming mobile intelligent applications, such as augmented reality (AR) games, three-dimension (3D) reconstruction, and automatic robotics. 
The advancements on machine learning (ML) provide a powerful tool for stable and reliable applications. 
As an important aspect of ML, deep neural network (DNN) inference is always computation-intensive, which makes it difficult for limited-resource devices to complete the execution process within an acceptable latency.
This difficulty can be usually solved in two ways, i.e., by either configuring a relevant model on the server with sufficient resource to execute the computation task or offloading the computation task from local devices to the cloud server. 
However, the former way cannot guarantee the most suitable model for each task, while the latter way will lead to additional transmission cost and privacy leakage issues. 
Thus, it is important to investigate the ML model splitting technology, which can split an ML model and deploy splitted computation tasks onto multiple edge devices with high efficiency and low latency. 

Model splitting framework can partition a DNN model into several parts. Each part is calculated by one device and the calculation result is passed to the corresponding device based on the model splitting framework via wireless or wired links.
The 5G cellular network is a native mobile edge computing structure suitable for ML model splitting. 
The model splitting framework can be presented in a chain or mesh topology, which includes cellular user equipment (UE), small base stations (SBSs), macro base stations (MBSs) equipped with mobile edge computing (MEC) servers, and cloud servers.

Recently, there are some works focusing on 
applying the ML model splitting on the mobile edge devices in a cellular network, which allows each edge device to compute part of the ML model \cite{kang2017neurosurgeon,hu2019dynamic,wang2021hivemind,teerapittayanon2017distributed,li2018edge}.
The authors in~\cite{kang2017neurosurgeon} proposed a single-split method, which can partition a DNN model into two parts for two-device system.
Furthermore,  Hu et al.~\cite{hu2019dynamic} utilized directed acyclic graph theory in the single-split method.
The works in \cite{kang2017neurosurgeon} and \cite{hu2019dynamic} are limited to two-device system.
Considering the general multi-split problem, the work in~\cite{wang2021hivemind} constructed a min-cost graph search problem. However, the search algorithm in \cite{wang2021hivemind} is limited to the linear network, which cannot be applied to the complex mesh cellular network.
Besides, Teerapittayanonet al.~\cite{teerapittayanon2017distributed} proposed to distribute the given DNN model across computing hierarchies for the purpose of reducing the communication data size. 
Moreover, Li et al.~\cite{li2018edge} utilized an early-exit mechanism to adjust the splitting model size to accelerate the model inference.
However, the existing works \cite{kang2017neurosurgeon,hu2019dynamic,wang2021hivemind,teerapittayanon2017distributed,li2018edge} did not consider customizing the neural network according to edge nodes' computation and communication abilities, 
which indicate that the results obtained in \cite{kang2017neurosurgeon,hu2019dynamic,wang2021hivemind,teerapittayanon2017distributed,li2018edge} cannot  guarantee the performance including accuracy and completion latency requirements over a mobile edge network.

Typically, split learning in mobile edge networks generally involves three unique properties:
\emph{(1) Multi-split}:
The existence of multiple devices in the cellular network requires the multi-split of the DNN model so as to each device compute part of the model. 
The number of feasible split solution increases exponentially with the numbers of DNN layers and edge devices.
Thus, the linear exhaustive search algorithm used in single-split problem is not practical to solve the multi-split problem.
\emph{(2) Multi-object}:
The solution of ML model splitting needs to satisfy multiple optimization goals, such as model accuracy, completion latency, and data privacy.
\emph{(3) Complicated topology}:
According to the practical scenario of 5G cellular networks, the edge network topology can be summarized into two forms: chain and mesh. 
Correspondingly, the DNN includes linear and non-linear forms. 

To realize the purpose of deploying split learning over a wireless MEC system, the DNN model splitting should be adapt to computation and communication abilities of edge nodes. 
However, the simplified model splitting method cannot guarantee the convergence and latency performance of DNN. As a result, deploying split learning over a wireless network calls for the iterative design of model generating and model splitting, where model generating means that the DNN model should be properly designed and chosen based on the given training accuracy and latency requirements. 
To solve the model generating problem, neural architecture search (NAS) can be used to automatically design the artificial neural networks. 
The previous work \cite{chen2018searching,cai2018proxylessnas,he2018amc} utilized NAS for model generating in a centralized manner, which cannot be directly applied to the distributed mobile edge networks. 
This motivates us to utilize the NAS method in solving the split learning problem over a wireless MEC system.

In this paper, based on the emerging technique of NAS, we propose the so-called joint model split and neural architecture search (JMSNAS) framework for deploying an ML model over the MEC. 
We formulate the multi-split problem by searching for a network achieving the accuracy and latency trade-off on a cellular network with given multi-split points. 
To our best knowledge, JMSNAS is the first practical framework that splits the ML model for MEC systems by using the NAS method. 
Our main contributions are list as follows: 
\begin{itemize}
    \item
    We propose a multi-split algorithm, which can assign the DNN sub-model for each computation node in a given network topology. 
    \item
    By using multi-objective regularizations, the multi-objective mechanism is integrated in the loss function to various practical requirements.
    \item
    The proposed JMSNAS makes full use of all the devices in the cellular network, which can be applied to different MEC topology networks including both chain and mesh.
    \item 
    Experiment results indicates that the proposed JMSNAS outperforms the State-of-the-art splitting method in terms of accuracy and latency. 
\end{itemize}

%% file: _txt/2_preliminaries.tex
\section{Preliminaries}
\begin{figure}[t]
    \centering
    \includegraphics[width=0.95\linewidth]{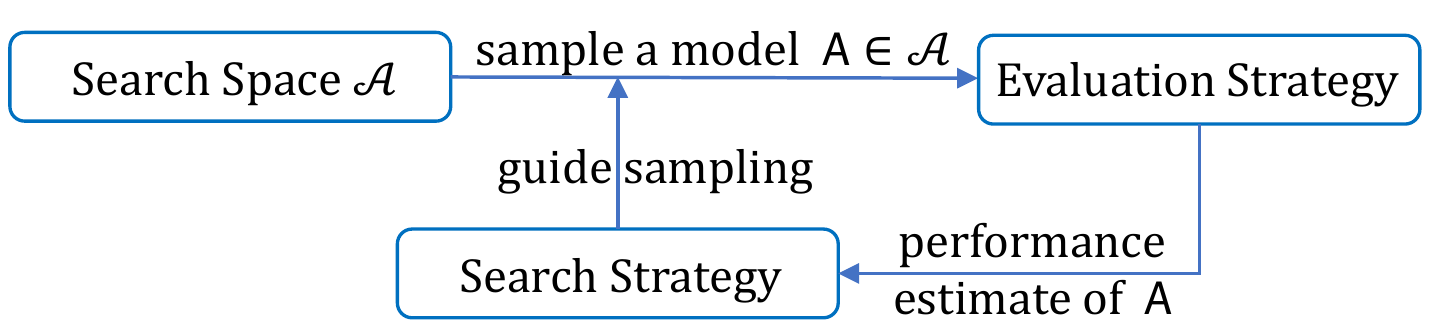}
    \caption{Three major components of NAS models.}
    \label{fig:overview of NAS}
    \vspace{-12pt}
    \end{figure}
\subsection{Neural Architecture Search}
NAS models have outperformed manually designed architectures in many tasks, such as semantic segmentation \cite{chen2018searching} and model compression \cite{he2018amc}.
The NAS process involves three major components: \textit{search space}, \textit{evaluation strategy}, and \textit{search strategy} (Fig.~\ref{fig:overview of NAS}).
Search space defines a family of candidate operations and the way operations connect. 
Evaluation strategy determines the quality metric of the candidate models to provide feedback that guides the search strategy.
Search strategy is the method to explore the search space and generate high-quality model architectures.

\subsection{ML Model Splitting over MEC Systems}
ML model splitting partitions the DNN computing load over MEC infrastructures, to meet specific requirements such as low inference latency. 
The typical scenarios of ML model splitting are shown in Fig. \ref{fig:senerio}.

Our NAS approach for ML model splitting considers both communication and computation costs of different devices in a cellular MEC network, and explore the comprehensive search space to obtain a DNN model with low latency and high accuracy on the given network.
Besides, our approach can naturally form a splitting scheme for the obtained DNN model that deploys different parts of the DNN to the devices in the target network. 
Finally, our approach can be applied on the general network structure, including both chain and mesh cellular networks.

%% file: _txt/3_Problem_Formulation.tex
\begin{figure}[t]
    \centering
    \subfigure[The chain MEC network]{   
    \begin{minipage}[b]{0.5\textwidth}   
    \includegraphics[width=0.95\textwidth]{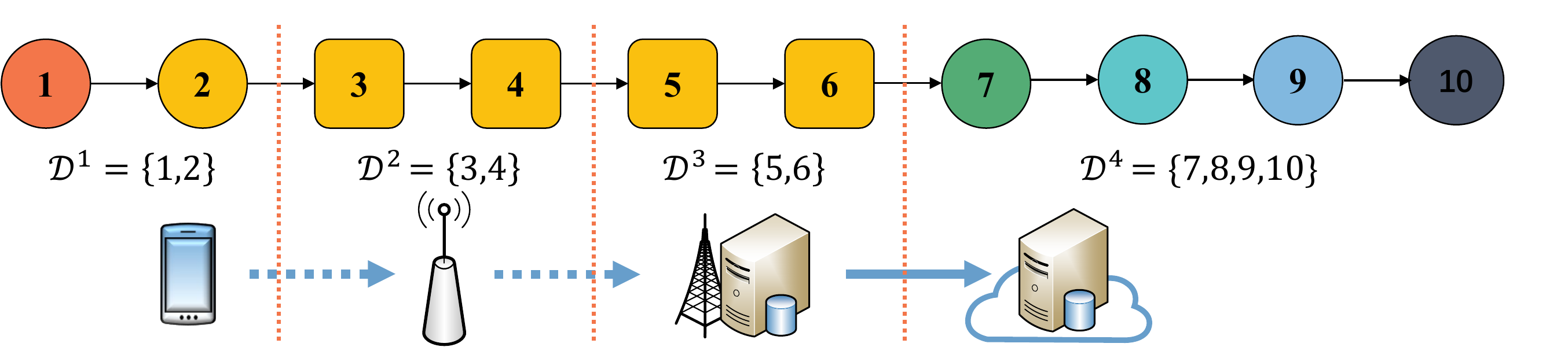}   \label{fig:chain}
    \end{minipage}}
    
    \subfigure[The mesh MEC network]{
    \begin{minipage}[b]{0.5\textwidth}
    \includegraphics[width=0.95\textwidth]{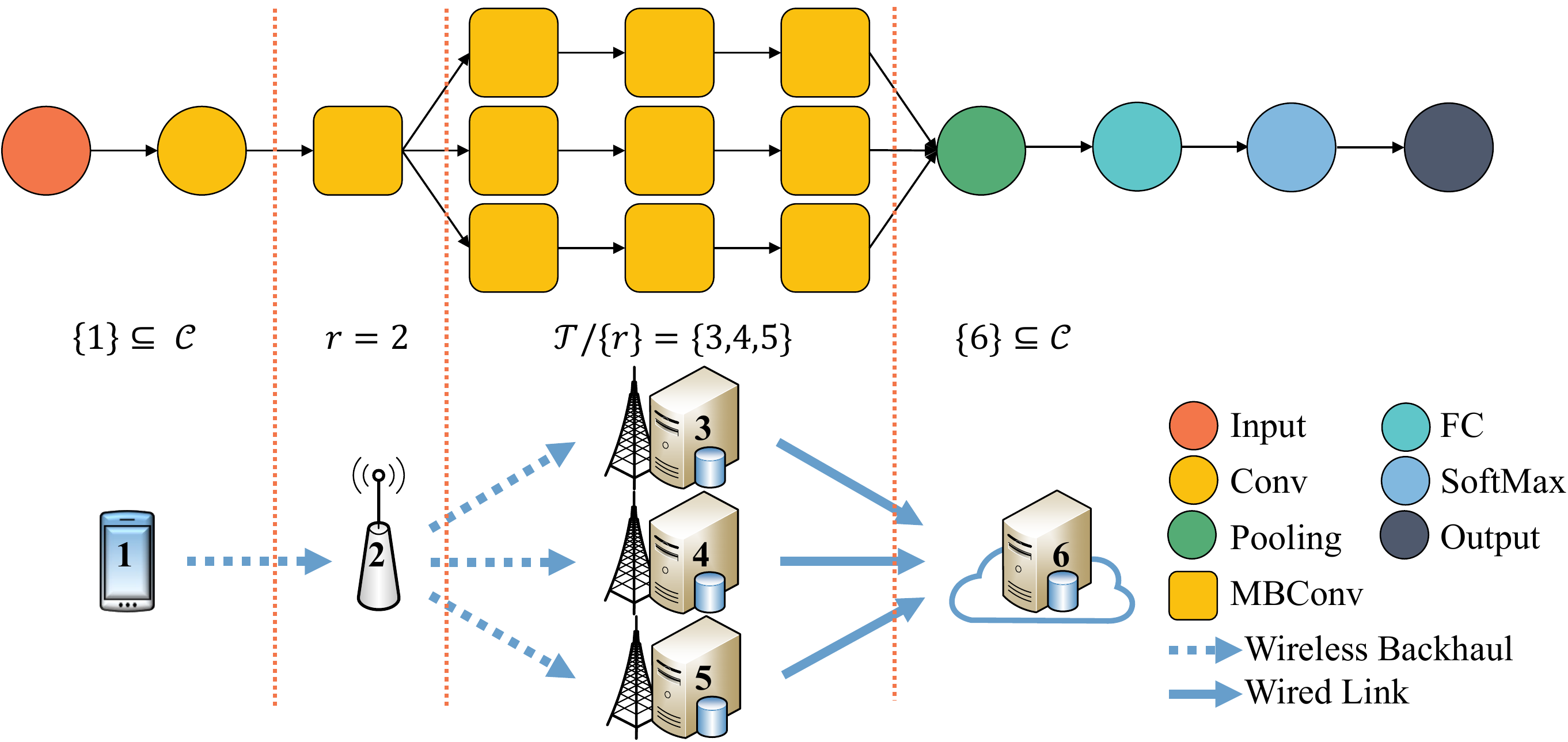}    \label{fig:mesh}
    \end{minipage}}
    
    \caption{The typical scenarios of ML model splitting.} \label{fig:senerio}
    
\end{figure}
\begin{figure*} [!t]
    \centering
    \includegraphics[width =0.95 \textwidth]{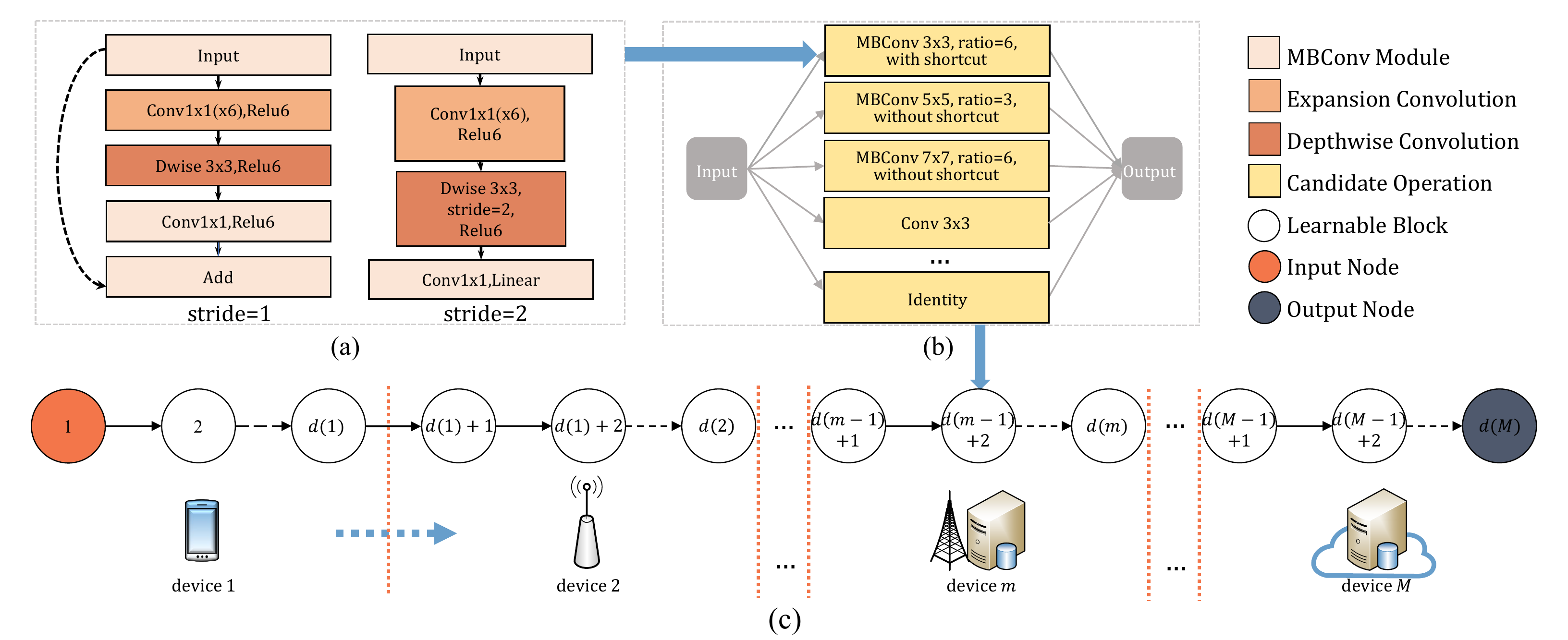}
    \caption{
    The implementation procedures. A Leader node collects the MEC network information and constructs the search space by selecting the modules in (a) and designing the candidate operations in (b). After running JMSNAS, the leader node sends the parts of the generated model to the corresponding devices in (c).
    }
    \vspace{-1em}
    \label{figure:framework}
    \end{figure*}

\section{System Model and Problem Formulation}
Consider a deep neural network, which can be described by a directed computational graph $G(e_1,\dots,e_N)$ as shown in Fig.~\ref{fig:senerio}.
The nodes in Fig.~\ref{fig:senerio} stand for the operations, while directed edges stand for data stream cross layers.
We use $[N]$ to represent set $\{1,\cdots,N\}$.
Denote $N$ and $M$ as the number of layers and available devices, respectively.
Layer $n \in [N]$ represents the $n$th node in the computational graph, and device $m \in [M]$ stands for the $m$th node in the MEC network.
Let $\tau_n^m$ be the execution latency of layer $n$ on device $m$ and $\varepsilon_n^m$ be the communication latency of transmitting the output of layer $n$ between device $m$ and the following device.
Specifically, we have $\varepsilon_{n}^M=0$ since device $M$ is the last device.
Both chain and mesh networks are considered in our model. 

\subsection{Chain Network}\label{chain_network}
As shown in Fig. \ref{fig:chain}, consider a network with one UE, $M-2$ edge devices, and a cloud server to form a $M$-node cellular chain. 
Since there are $M$ computation nodes, we need to split the original DNN network into $M$ parts and each node can compute one part of the DNN network.
Let $\mathcal{D}^m $ stand for the layers deployed on device $m$, where $\mathcal{D}^1=\{1,2\}$ means that both layers 1 and 2 are assigned to device 1.

With the above notation, the completion latency on chain network can be described as
\begin{equation}\small
    T = \sum_{m \in [M]} (\sum_{n\in \mathcal{D}^m} \tau_n^m + \varepsilon_{d(m)}^m),
    \label{chain}
\end{equation}
where $d(m) = \max{\mathcal{D}^m}$ denotes the index of the last layer executed on device $m$.

\subsection{Mesh Network}
Consider a mesh network where a UE first accesses an SBS through the wireless backhaul, then the SBS broadcasts the results to three MBSs, and finally the outputs of MBSs are aggregated in a cloud server, as shown in Fig.~\ref{fig:mesh}. 
Let $\mathcal{C}$ denote the set of devices connected in the chain
and $\mathcal{T}$ denote devices connected in the tree form. For example, in Fig.~\ref{fig:mesh}, $\mathcal{C} = \{1,6\}$, $\mathcal{T} = \{2,3,4,5\}$. 
There is a root node $r$ in tree nodes set such as device 2 in Fig.~\ref{fig:mesh}, which forwards the output data of layer $d(2)$ to devices in set $\mathcal{T}\backslash\{r\}$ for executing the following layers in parallel.
Thus, the completion latency on mesh network can be described as
\begin{equation}\small
    \begin{aligned}
    T&=\sum_{m \in \mathcal{C}} (\sum_{n\in \mathcal{D}^m} \tau_n^m + \varepsilon_{d(m)}^m) \\
    &+\max_{m \in \mathcal{T}\backslash\{r\}} (\sum_{n\in \mathcal{D}^r} \tau_n^r + \varepsilon_{d(r)}^r+ \sum_{n\in \mathcal{D}^m} \tau_n^m + \varepsilon_{d(m)}^m).
    \end{aligned}
    \label{mesh}
\end{equation}

\subsection{Implementation Model}
The detailed implementation procedures of JMSNAS mainly include three steps. 
In the first step, a leader node collects the device information in the cellular edge network, including the computing capabilities of each device, the communication capabilities between devices, and the device connection topology. 
Having obtained the device information, the leader node in the second step completes the initialization of the NAS search space, i.e., the connection mode of the DNN layers and the maximum number of DNN layers which can be executed by each device. 
In other words, the structure of computational graph (Fig.~\ref{figure:framework}(c)) and parameters $D^m$, $d(m)$ are determined for each $m$ in the second step.
According to the characteristics and complexity of the DNN task, the candidate operations in the search space are artificially set.
In the third step, the leader node runs JMSNAS to determine the DNN used for the specific task. 
Additionally, the network naturally has a split matching scheme deployed to each device to meet the constraints set by the task, which can include limited latency, limited power consumption, and so on.
The leader node sends the parts of searched network to the corresponding device, and each device in the MEC network executes the computation and communication assignment in a distributed manner.


\begin{figure*} [!htb]
    \centering
    \includegraphics[width = 0.95\textwidth]{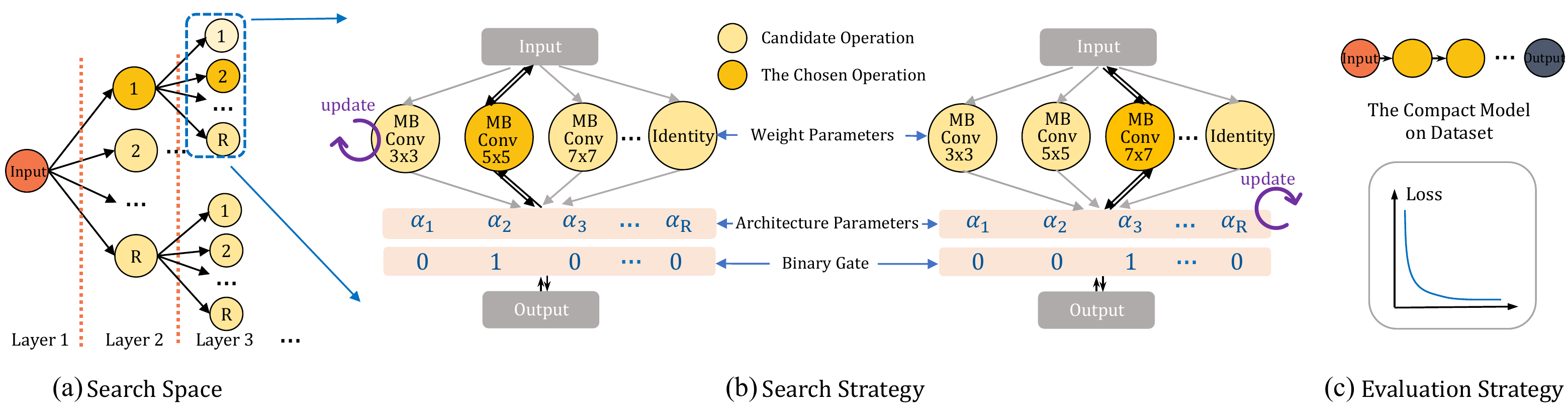}
    \caption{The overview of the NAS method. Search space is a full-tree structure (a) making up of all the possible DNN models. 
    The relative parameters are updated in (b), and evaluated in (c).  }
    \vspace{-1em}
    \label{figure:update}
    \end{figure*}

\subsection{Problem Formulation} \label{sec:Optimization Problem}

Our goal is to optimize the neural network weight parameter $\bm{\theta}$ and architecture parameter $\bm{\alpha}$ so as to minimize the loss function of the DNN model under given latency constraint. 
Mathematically, the considered optimization problem can be formulated as:
\begin{equation}\small
\label{eq:loss}
    \begin{aligned}
    \min~&L = \sum_{k=1}^K \ell(\bm{\alpha},\bm{\theta};(\bm{x}_k,y_k)), \\
    s.t.~&T \leq T_{Const},
    \end{aligned}
\end{equation}
where $\bm{\alpha}=[\alpha^1_1,\alpha^1_2,\cdots,\alpha^n_i,\cdots,\alpha^N_R]$ and $T_{Const}$ is the maximum allowed latency. The variable $\alpha^n_i, n\in[N],i\in[R]$ indicates the weight of choosing the $i$th candidate as the operation of layer $n$.
In \eqref{eq:loss}, $\bm {x_k}$ and $y_k$ respectively indicate the $k$th sampled input image and the corresponding label, $S$ is the number of sampled images. 
$\ell(\cdot)$ represents the cross entropy loss function.
Due to complicated non-convex loss function and latency constraint, it is generally difficulty to solve problem \eqref{eq:loss} with the conventional convex optimization theory. 
To solve it, we use the advancement of ML in the following section. 

%% file: _txt/4_Methodology.tex
\section{Methodology}
In this section, we first introduce the JMSNAS framework, which includes search spaces, evaluation strategies, and search strategies for optimized neural networks \cite{cai2018proxylessnas}. Then, we present a gradient-based algorithm to deal with the non-differential device metric: latency.
\subsection{Neural Architecture Search} \label{sec:NAS}
Through NAS techniques, we can automatically develop a model that outperforms previous designs deployed on cellular MEC networks. 

\vspace{0.4em}\noindent \textbf{Search Space:} 
Let $\mathcal{V}_n$ be the set of $R$ candidate operations available for layer $n$.
In the initialized computational graph with fixed topology, the candidate operations of each layer together constitute a full-tree search space, as shown in Fig.~\ref{figure:update}(a).
Our search space involves all combinations of $R$ operations for $N$ layers,
which is sufficient to specific task with adjustable parameters $N$ and $R$.

As an example to construct search space $\mathcal{V}_n$, 
the MobileNetV2 is a lightweight and highly efficient model \cite{sandler2018mobilenetv2}, which performs well on large-scale image classification tasks.
It is constructed by module mobile inverted bottleneck convolution (MBConv), which can be used as the backbone to build the candidate operations. 
The module MBConv takes low-dimension vectors as input, expands to high dimension, and is then filtered with a depthwise convolution. 
With changeable parameters about expansion ratio and convolution kernel size, $\mathcal{V}_n$ includes the following types:
    \begin{itemize}
        \item $3 \times 3$ MBConv with expansion ratio 3
        \item $3 \times 3$ MBConv with expansion ratio 6 
        \item $5 \times 5$ MBConv with expansion ratio 3
        \item $5 \times 5$ MBConv with expansion ratio 6  
        \item $7 \times 7$ MBConv with expansion ratio 3 
        \item $7 \times 7$ MBConv with expansion ratio 6 
        \item Identity
    \end{itemize}
In addition, each operation can choose whether to have a shortcut connection or not. Therefore, there are up to 13 optional operations. 
The network length can be shortened by selecting identity to skip blocks.
To make the model more accurate, the framework might choose a large kernel and a high expansion ratio with a large amount of computation, which leads to a larger network.
On the contrary, to save execution latency, the framework might choose a small kernel and a low expansion ratio, which makes the network thinner. 
As a result, in order to balance the accuracy and latency, the width and length of the model should be well designed. 

\vspace{0.4em}\noindent \textbf{Evaluation Strategy:} 
Before demonstrating the evaluation strategy, we need to clarify how to represent the forward propagation result of the full-tree structure as shown in Fig.~\ref{figure:update}(a).
To construct the full-tree structure that includes all the combinations in the search space, we denote the operation of each layer in the computational graph by $\bm v=[v_1,\cdots,v_R]^T$, which is a mixed operation vector with $R$ elements. The output of $\bm{v}$ is designed based on the output of its $R$ paths. 

To simplify the description, we use a certain DNN layer to illustrate our design for mixed operations. 
Without loss of generality, we replace $\alpha_{i}^{n}$ with $\alpha_{i}$ in the real-valued architecture parameters.
Moreover, we introduce the one-hot binary gate $\bm g = [g_1,\cdots,g_R]^{T}$, where $g_i=1$ with probability $p_i$,
and $p_i = {\exp \left(\alpha_{i}\right)}/{\sum_{j} \exp \left(\alpha_{j}\right)}$  indicates the probability of choosing operation $i$. 
With input $\bm {x}_k$, the output of $\bm v$ can be formulated as $\sum_{i=1}^{R} g_{i} v_{i}(\bm {x}_k) = \bm g^T \bm v(\bm {x}_k)$, where $v(\bm {x}_k)=[v_1(\bm {x}_k),\cdots,v_R(\bm {x}_k)]^T$.

To solve problem \eqref{eq:loss}, we modify the objective function as
\begin{equation}\small
    \label{Loss}
    \sum_{k=1}^K \ell(\bm{\alpha},\bm{\theta};(\bm {x}_k,y_k)) 
    + \lambda_1 ||\bm{\theta}||_2^2
    + \lambda_2 {(T-T_{Const})}^2,
\end{equation}
where $\lambda_1, \lambda_2$ are hyper-parameters to adjust the learning process. 
Note that the latency constraint is formulated as a penalty in \eqref{Loss}, which forces that the optimal solution satisfying $T=T_{const}$. This is because the optimal solution of  \eqref{eq:loss} is always achieved at $T=T_{const}$ as deep and time-consuming network can lead to small loss value.


\vspace{0.4em}\noindent \textbf{Search Strategy:} 
There are two types of parameters in our framework, i.e., weight parameter $\bm{\theta}$ and architecture parameter $\bm{\alpha}$.
We train weight and architecture parameters in an alternating manner, as shown in Fig. \ref{figure:update}(b).
When training weight parameters, the architecture parameters are fixed and the binary gates $\bm g$ are sampled to identify the current DNN model. 
Then,  the sampled model is trained with forward and backward propagation. 
When updating architecture parameters, the weight parameters are given in the previous step and the binary gates are reset.
To update the  architecture parameter,  the partial derivative $ {\partial L}/{\partial \alpha_{i}}$ with respect to discrete operation choosing needs to be calculated, which is provided by the following lemma.   
\begin{lemma}
The  partial derivative  ${\partial L}/{\partial \alpha_{i}}$ can be approximately presented by  
\begin{equation}\small\label{lemma1der}
\sum_{j=1}^{R} \frac{\partial L}{\partial g_{j}} p_{j}\left(\delta_{i j}-p_{i}\right),
\end{equation}
where $\delta_{ii}=1$ if $i=j$, $\delta_{ij}=0$ if $i\neq j$, and $\frac{\partial L}{\partial g_{j}}$ can be obtained from the following equation \eqref{ce-der}.
 
\end{lemma} 
\itshape {Proof:} \upshape
The partial derivative of $L$ with respect to $\alpha_i$ is: 
\begin{equation}\small
    \begin{aligned}
        \frac{\partial L}{\partial \alpha_{i}}&=\sum_{j=1}^{R} \frac{\partial L}{\partial p_{j}} \frac{\partial p_{j}}{\partial \alpha_{i}} \approx \sum_{j=1}^{R} \frac{\partial L}{\partial g_{j}} \frac{\partial p_{j}}{\partial \alpha_{i}}\\
        &=\sum_{j=1}^{R} \frac{\partial L}{\partial g_{j}} \frac{\partial\left(\frac{\exp \left(\alpha_{j}\right)}{\sum_{k} \exp \left(\alpha_{k}\right)}\right)}{\partial \alpha_{i}}=\sum_{j=1}^{R} \frac{\partial L}{\partial g_{j}} p_{j}\left(\delta_{i j}-p_{i}\right),
    \end{aligned}
    \label{loss}
\end{equation}
where the first equality follows from the chain theory and the approximation holds based on the definition of $g_j$.
The derivative $\partial L /\partial g_{j}$ can be calculated by substituting the expression of $g_{j}$ into the function \eqref{Loss}. 
In particular, if we consider the cross-entropy loss function, equation \eqref{Loss} can be further rewritten as
\begin{equation}\small
    \begin{aligned}
    L_{\text{CE}}=-\frac{1}{K} \sum_{k=1}^{K} (&y_k \log h_{\bm g}\bm v(\bm {x}_k) 
    \\ & +(1-y_{k})\log (1-h_{\bm g}\bm v (\bm {x}_k))),
    \end{aligned}
\end{equation}
where $h_{\bm g}\bm v (\bm {x}_k) = 1/(1+e^{-\bm{g}^T\bm v(\bm {x}_k)})$ indicates the predicted probability. 
Thus, we can obtain
\begin{equation}\small\label{ce-der}
    \frac{\partial L_{\text{CE}}}{\partial \bm g}=\frac{1}{K} \sum_{k=1}^{K}\left(h_{\bm g}\bm v (\bm {x}_k)-y_{k}\right) \bm v(\bm {x}_k).
\end{equation}
This completes the proof. 
\hfill $\Box$


Based on Lemma 1, we can update the architecture parameter through backpropagation.
The architecture parameter updating procedure involves calculating and storing $v_j(\bm x)$ for every $j$, which costs $R$ times memory.
To address this issue, we mask all the paths except for the sampled two in every training process so that we can reduce the memory cost from $R$ times to 2 times.
\subsection{Derivative of  Latency Function} \label{sec:Make Latency Differentiable}
Since our training network dynamically chooses operations according to a probability distribution, the latency in loss function \eqref{Loss} is not differentiable with respect to the architecture parameter.
To handle this issue, we reformulate the latency of a network to the average latency, which is a continuous function.
A mixed operation $\bm v$ includes candidate set $\{v_1, v_2, \cdots, v_R\}$ and each operation $v_i$ corresponds to a selection probability $p_i$.
We build a regression model $U(\cdot)$ to estimate the operation latency. For example, when layer $n$ adopts operation $v_i$ executed on device $m$, we have $U_n^m(v_i)$ as the execution latency.  
In such a full-tree structure, the execution latency of layer $n$ on device $m$, $\tau_n^m$ in \eqref{chain} and \eqref{mesh} should be reformulated as   
$\mathbb{E}(\tau_n^m)=\sum_{i}{p_i U_n^m(v_i)}$.
Thus, the gradient of $\mathbb{E}(\tau_n^m)$ with respect to architecture parameter can be given by: ${\partial \mathbb{E}(\tau_n^m)}/{\partial p_{i}}$.
Furthermore, we represent the overall latency $\mathbb{E}_{\bm{\alpha}}(T)$ by replacing $\tau_n^m$ with $\mathbb{E}(\tau_n^m)$.

In summary, the proposed method provides ample search space and sufficient flexibility to search for proper layer operations and enables high performance as the trade-off between accuracy and latency. 
The cost during the NAS process is completely undertaken by the leader node, which requires the leader node to have strong computing capabilities. 

%% file: _txt/5_experiment.tex
\begin{table}[t]
  \footnotesize
  \centering
   \renewcommand{\arraystretch}{1.1}
  \caption{Link settings}
  \label{tab:Link settings}
   \resizebox{0.75\linewidth}{!}{%
  \begin{tabular}{| c | c | c | c |}
      \hline
      Transmitter & Receiver & Type & Capacity(Mbps)\\
      \hline 
      UE &  SBS & Wireless & 25\\ 
      \hline 
      SBS &  MBS & Wireless & 50\\ 
      \hline 
      MBS &  Cloud & Wired & 200\\ 
	  \hline
  \end{tabular}}
  \vspace{-2pt}
\end{table}
\section{Experiment Results} 
In this section, we first describe our experiment setup, including the cellular network setup, and the configuration for the NAS procedure. We then present our framework evaluation results on ML model splitting compared with the previous methods.  
\subsection{Cellular Network} 
The performance of the proposed JMSNAS framework is evaluated on both chain \ref{fig:chain} and mesh \ref{fig:mesh} networks.
Table \ref{tab:Link settings} shows the communication link settings between devices.

The execution and communication latency profiles of different operations involved in the DNN model are the key metric of the model workload.
To accurately measure the operation latency, we adopt a Pytorch~\cite{pytorch} package (torchprof) to track latency on different devices for each involved operation and build an estimator $U(\cdot)$ to predict operation-wise latency during model inference.

We measure the latency profiles on four types of machines to represent the UE, SBS, MBS with MEC server and cloud server, respectively: \\
(1) Raspberry Pi 4 Model with 4 Cortex-A72 1.5GHz CPUs,\\
(2) XPS15 Laptop with Intel i7-11800H CPU, 16GB DDR 4 RAM, and Nvidia RTX3050 GPU \\
(3) NVIDIA Jetson AGX Xavier with 64 Tensor Core GPUs and 8-core ARM CPUs, \\
(4) A server with two Intel Platinum 8280 CPUs, and Tesla V100 GPU. 

\subsection{Neural Architecture Search} 
We demonstrate the effectiveness of JMSNAS on the ImageNet dataset~\cite{ILSVRC15}.
The training set includes 1231167 images of 1000 classes, each with dimension of $224\times224\times3$, while the validation set includes 50000 images.

In order to obtain a model that performs well on the given cellular network, the NAS process consists of two stages. 
In the first stage, we search the full-tree structure on the training split for 20 warm-up epochs and 60 training epochs, using Adam optimizer with initial learning rate of 0.002 and batch size of 512.
Warm-up training is a technique widely used in deep learning. It helps to alleviate overfitting on the mini-batchs, and to maintain the stability of the DNN model.
At the end of every training epoch, we evaluate the performance of the current compact network on the validation set.
We set up three levels of completion latency constraints on the chain network and mesh network, respectively.

In the second stage, after the architecture parameters of the full-tree structure converge, the compact model architecture is fixed. 
Then, we further train the model on the training set for 200 epochs, with the weight parameters in the first stage as the pretrained parameters.
In this way, the derived compact model will get higher accuracy without compromising its efficiency.

\begin{figure}[t]
  \centering
  \includegraphics[width=0.91\linewidth]{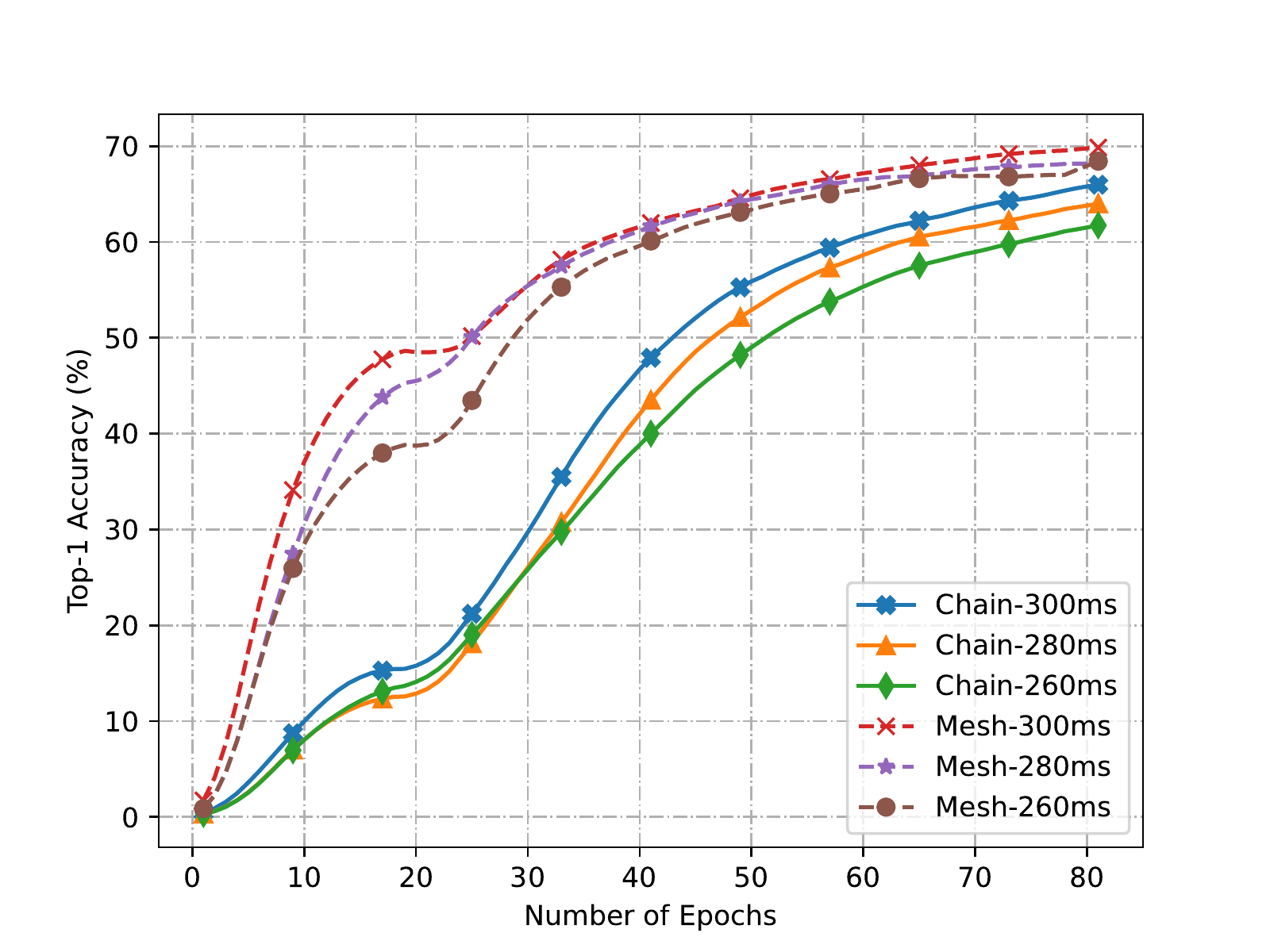}
  \caption{In the first NAS stage, JMSNAS on the chain and mesh networks with different $T_{Const}$ latency constraints.}
  \label{fig:accuracy}
  \vspace{-12pt}
  \end{figure}

We adopt two performance metrics of our framework, i.e., top-1 accuracy and latency.
Fig. \ref{fig:accuracy} shows the top-1 accuracy corresponding to different latency constraints in the first NAS stage.
Here, we find that the models did not reach the highest accuracy during the first stage, since the model architecture is still alternating.
And when the maximum allowed latency is larger, the model accuracy tends to be higher.
Compared with models on the chain network, models on the mesh network shows faster convergence and higher accuracy with the same number of training epochs. 

Since the operations in our search space are mainly MBConvs, modules in MobileNetV2, we also compare the accuracy and latency of the JMSNAS-crafted models with MobileNet (V1\cite{howard2017mobilenets}, V2\cite{sandler2018mobilenetv2}, V3\cite{howard2019searching}).
Fig.~\ref{fig:compare} shows that our framework achieves higher accuracy, which confirms the effectiveness of JMSNAS.
As for the latency, we compare JMSNAS-crafted model performance with two baselines, cloud computing and HiveMind \cite{wang2021hivemind} multi-split framework.
We apply the cloud computing to the chain model obtained by JMSNAS, by uploading the input data to the cloud center and completing the model inference on the cloud. 
The latency is mainly determined by the communication link conditions.
Fig.~\ref{fig:compare} shows that JMSNAS can reduce the average latency by up to 20.1\% compared with cloud computing.
For DNN models with a chain structure such as MobileNet series, we can adopt linear search method like HiveMind to determine the best splitting points.
Fig. \ref{fig:compare} shows that our framework outperforms MobileNet with linear search method.

%% file: _txt/6_conclusion.tex
\begin{figure}[t]
  \centering
  \includegraphics[width=0.95\linewidth]{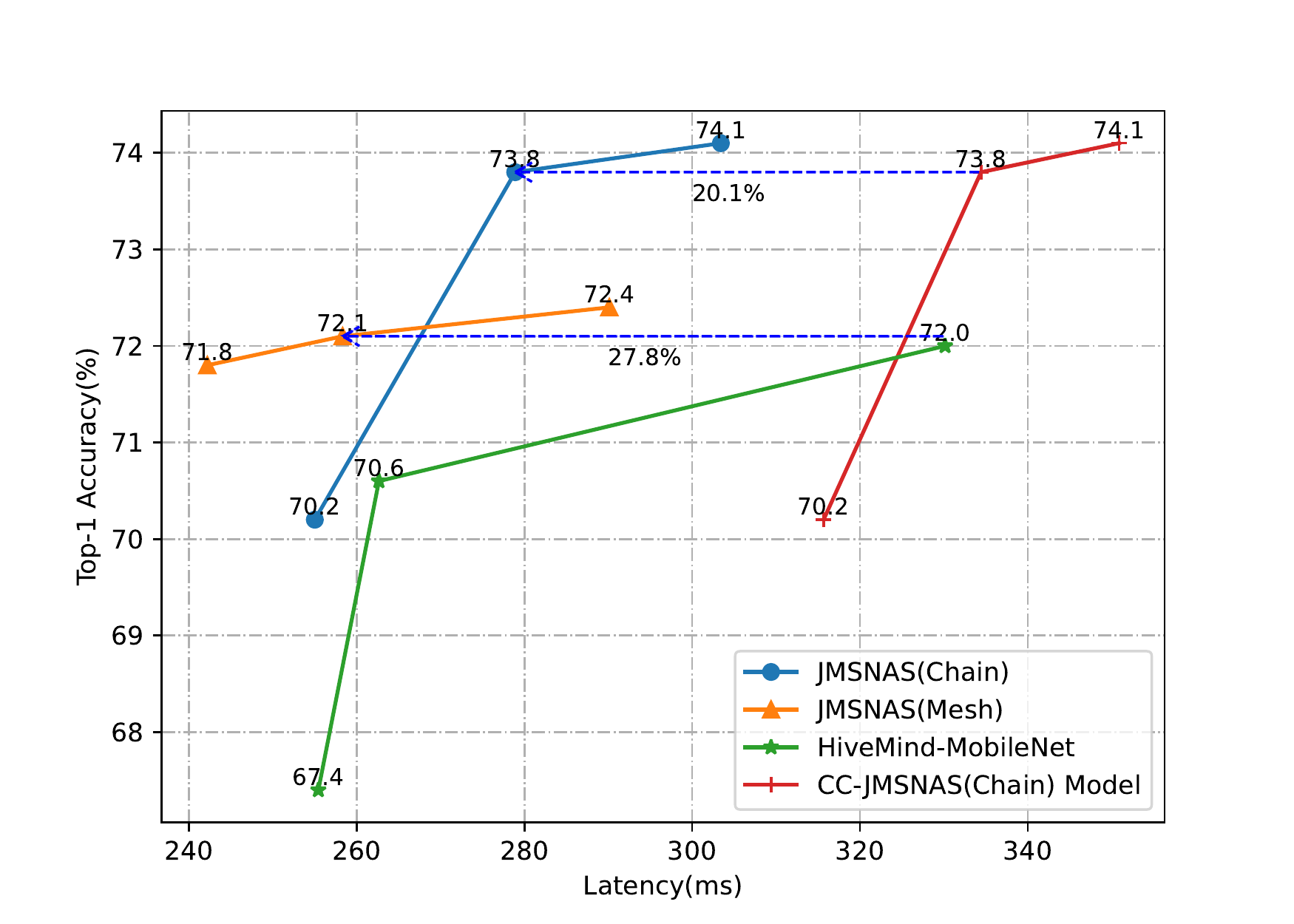}
  \caption{Final accuracy vs. latency after the second NAS stage.}
  \label{fig:compare}
  \vspace{-12pt}
  \end{figure}
\section{Conclusion}
In this paper, we have proposed a NAS-based multi-split framework, deploying the generated DNN model to a cellular mobile edge network to meet the accuracy and latency requirements.
The proposed JMSNAS works well on large-scale image classification tasks with an ample search space dependent on the mobile edge network conditions.
The automatically generated models with native split scheme outperform the previous model split method. 

The proposed JMSNAS is a general framework, and it can be applied in any practical scenario to dynamically determine a customized DNN.  

